\documentclass[runningheads]{llncs}
\usepackage[T1]{fontenc}
\usepackage{graphicx}
\usepackage{tabularx}
\usepackage{array}
\usepackage{CJKutf8} 
\begin{document}
\begin{CJK*}{UTF8}{gbsn}
\title{Automatic Evaluation Metrics for Document-level Translation: Overview, Challenges and Trends}
\titlerunning{Automatic Evaluation Metrics for Document-level Translation}
%
\author{Jiaxin GUO, Xiaoyu Chen, Zhiqiang Rao, Jinlong Yang, Zongyao Li, \\Hengchao Shang, Daimeng Wei and Hao Yang}
\authorrunning{Jiaxin GUO, Xiaoyu Chen et al.}
%
\institute{Huawei Translation Services Center, Beijing, China\\
\email{\{jiaxinguo1,chenxiaoyu35,weidaimeng,yanghao30\}@huawei.com}}
\maketitle              
\begin{abstract}
With the rapid development of deep learning technologies, the field of machine translation has witnessed significant progress, especially with the advent of large language models (LLMs) that have greatly propelled the advancement of document-level translation. However, accurately evaluating the quality of document-level translation remains an urgent issue. This paper first introduces the development status of document-level translation and the importance of evaluation, highlighting the crucial role of automatic evaluation metrics in reflecting translation quality and guiding the improvement of translation systems. It then provides a detailed analysis of the current state of automatic evaluation schemes and metrics, including evaluation methods with and without reference texts, as well as traditional metrics, Model-based metrics and LLM-based metrics. Subsequently, the paper explores the challenges faced by current evaluation methods, such as the lack of reference diversity, dependence on sentence-level alignment information, and the bias, inaccuracy, and lack of interpretability of the LLM-as-a-judge method. Finally, the paper looks ahead to the future trends in evaluation methods, including the development of more user-friendly document-level evaluation methods and more robust LLM-as-a-judge methods, and proposes possible research directions, such as reducing the dependency on sentence-level information, introducing multi-level and multi-granular evaluation approaches, and training models specifically for machine translation evaluation. This study aims to provide a comprehensive analysis of automatic evaluation for document-level translation and offer insights into future developments.

\end{abstract}
\section{Introduction}
\subsection{The Development of Document-level Translation}

In recent years, the rapid development of deep learning has brought unprecedented breakthroughs to the field of machine translation \cite{DBLP:conf/nips/VaswaniSPUJGKP17,sennrich2016improving,wei2023text,DBLP:conf/iclr/Gu0XLS18,DBLP:conf/emnlp/GhazvininejadLL19,DBLP:conf/inlg/WangGWCSWZTY21,DBLP:journals/corr/abs-2112-11640,DBLP:journals/corr/abs-2112-11642}. Early research mainly focused on improving the quality of sentence-level translation. By optimizing algorithms and model architectures, machine translation systems achieved significant improvements in accuracy and fluency at the sentence level. However, as application scenarios continue to expand, the demand for document-level translation has been growing rapidly. This requires machine translation systems to handle longer and more complex text structures while maintaining overall semantic coherence and consistency.

Recently, the emergence of large language models (LLMs) \cite{DBLP:journals/corr/abs-2309-16609,DBLP:journals/corr/abs-2303-08774,DBLP:journals/corr/abs-2302-13971,DBLP:journals/corr/abs-2307-09288,DBLP:journals/corr/abs-2407-10671} has brought new hope to document-level translation \cite{DBLP:conf/discomt/KimTN19,DBLP:journals/csur/MarufSH21,DBLP:conf/acl/FernandesYNM20,DBLP:conf/acl/CuiDZX24,DBLP:conf/wmt/WuH23,DBLP:journals/corr/abs-2401-06468,DBLP:conf/emnlp/WangTWL17,DBLP:conf/emnlp/TanZZ21,DBLP:conf/emnlp/LyuLGZ21}. Leveraging their powerful language generation capabilities and deep understanding of context, LLMs can produce more natural, fluent, and semantically coherent translation texts. These models have demonstrated significant advantages in handling long texts, leading to substantial improvements in the quality of document-level translation. For example, some translation systems based on LLMs can better capture logical relationships and thematic consistency within the text, resulting in higher-quality translation outcomes.

\subsection{The Importance of Evaluation}

Despite the significant progress made by LLMs in document-level translation, how to accurately evaluate the quality of document-level translation remains an urgent issue. Accurate evaluation is essential for understanding the performance of current machine translation systems and providing direction for future research and development. In practical applications, the quality evaluation of document-level translation is crucial for ensuring the accuracy and reliability of translations, especially in professional fields such as legal, medical, and technical translation tasks.

The most accurate evaluation method is undoubtedly human evaluation. Human translators, with their language knowledge and understanding of the text, can conduct detailed and nuanced evaluations of translation quality. However, human evaluation has many limitations. First, it is time-consuming and labor-intensive, making it cost-prohibitive for large-scale translation tasks. Second, human evaluation is highly subjective, with significant variations among different evaluators. Moreover, the speed of human evaluation is relatively slow, making it difficult to meet the demands of real-time and large-scale translation needs.

Therefore, the development of an accurate and effective automatic evaluation metric has become extremely important. An ideal automatic evaluation metric should be able to quickly and accurately assess translation quality and provide interpretable evaluation results. Such a metric can not only better reflect the performance of current machine translation systems but also provide valuable feedback for system improvement. In addition, automatic evaluation metrics can be used to guide the training and optimization of machine translation systems. An accurate and effective evaluation metric can provide more accurate feedback to the model, thereby improving the training effectiveness and final performance of the model.

In summary, with the increasing demand for document-level translation, the development of more rational and accurate automatic evaluation metrics is of great significance for advancing machine translation technology. This not only helps improve translation quality but also provides direction for further improvement of translation systems.

\section{Current Status}

\subsection{Evaluation Schemes}
Automatic evaluation schemes are mainly divided into two scenarios: with reference and without reference.

\subsubsection{With Reference} In this case, evaluation is based on the original text by comparing the similarity between the translation result and the reference. This method relies on high-quality reference translations, and the similarity between the translation result and the reference translation is used to measure the accuracy of the translation.

\subsubsection{Without Reference} In this case, evaluation is directly targeted at the consistency between the translation result and the original text. Since there is no reference translation for comparison, this method usually needs to assess whether the translation result can accurately convey the meaning of the original text while maintaining the fluency and naturalness of the language.

\subsection{Evaluation Metrics}

Evaluation metrics are mainly divided into three categories: Traditional Metrics, Model-based Metrics and LLM-based Metrics.

\subsubsection{Traditional Metrics} BLEU \cite{DBLP:conf/acl/PapineniRWZ02} is a widely used metric that mainly assesses the n-gram similarity between the translation result and the reference translation. It was initially designed for sentence-level machine translation evaluation but can also be directly extended to document-level translation evaluation. The main advantage of BLEU is its simple calculation and fast speed, but it mainly focuses on lexical matching and has a weaker ability to understand semantics.

\subsubsection{Model-based Metrics} Model-based metrics are a class of evaluation metrics that leverage pre-trained language models to assess translation quality. These metrics, including Comet \cite{DBLP:conf/emnlp/ReiSFL20} and BertScore \cite{DBLP:journals/corr/abs-1904-09675}, are trained on pre-trained models such as BERT \cite{DBLP:journals/corr/abs-1810-04805} and RoBERTa \cite{DBLP:journals/corr/abs-1907-11692}, which endow them with the ability to capture deeper semantic information compared to traditional metrics.

Comet is a model-based metric that can reflect semantic similarity. Compared with metrics like BLEU, Comet can better capture the semantic relationship between the translation result and the reference translation. Comet includes several variants, such as Comet20 and Comet22 for evaluation with reference, and CometKiwi for evaluation without reference. In sentence-level translation, Comet is considered to have the best consistency with human evaluation. 

To extend the application of these metrics to the document level, \cite{vernikos2022embarrassinglyeasydocumentlevelmt} proposed an innovative method that involves encoding additional context information. This approach, which we refer to as \textit{d}-Comet, allows the metric to consider the broader context of the document, thereby providing a more comprehensive evaluation of translation quality at the document level. By incorporating context, \textit{d}-Comet can better assess the overall coherence and consistency of the translation within the document, moving beyond the limitations of sentence-level evaluation.

\subsubsection{LLM-based Metrics} With the development of LLMs, their capabilities have become increasingly powerful, and the LLM-as-a-judge \cite{gu2025surveyllmasajudge} evaluation method has been applied in more and more natural language processing (NLP) fields. In this method, LLMs are used to perform scoring, ranking, or selection in various tasks and applications.

In some machine translation research, the LLM-as-a-judge evaluation method has also been applied to the evaluation of document-level translation. Usually, we use particularly powerful models, such as GPT-4 \footnote{https://openai.com/} or DeepSeek \footnote{https://www.deepseek.com/}, to conduct relative evaluations of translation results. Since the output of LLMs is stochastic, multiple evaluations of the same test set are usually required in practice, and the average value is then calculated. In addition, in relative evaluations, different results need to be randomly shuffled to mitigate the LLM's preference for different positional information.

\section{Challenges}

\subsection{Lack of Reference Diversity}

The diversity of reference translations is a key issue in machine translation evaluation. Especially for document-level translations generated by LLMs, the diversity is extremely high, and a single reference translation is insufficient to accurately measure translation quality. This is because LLMs can generate a variety of different but semantically equivalent translations, and a single reference translation may not cover all possible correct translations. Although generating diverse reference translations could solve this problem, the cost is very high, requiring a significant amount of time, resources, and professional human effort for annotation and validation.

\subsection{Sentence-level Alignment Dependency of \textit{d}-Comet}

\textit{d}-Comet provides a theoretical solution for extending sentence-level evaluation metrics to the document level, but it faces many challenges in practical application. \textit{d}-Comet strongly relies on the original text, reference translation, and machine translation output being segmented into an equal number of sentences and precisely aligned at the sentence level. However, translations generated by LLMs may creatively combine multiple consecutive sentences from the original text into a single sentence, making sentence-level alignment very difficult. Additionally, segmenting document-level translations into sentence-level units is also a challenge, especially for long texts. These issues make it difficult to effectively use \textit{d}-Comet in practical evaluations.

\subsection{Bias, Inaccuracy, and Lack of Interpretability of LLM-as-a-judge}

Despite the widespread application of the LLM-as-a-judge evaluation method in natural language processing (NLP), it has many problems in practical use.

\subsubsection{Bias}
Studies \cite{wataoka2024selfpreferencebiasllmasajudge} have shown that the LLM-as-a-judge method has a strong bias. For example, for a given document, if translations are generated by both GPT-4 and DeepSeek, and these translations are evaluated by GPT-4 and DeepSeek respectively, they will both tend to consider their own generated results as better. However, human evaluation might find these two results to be comparable.

\subsubsection{Inaccuracy}
The LLM-as-a-judge method is highly inaccurate in some cases. In our experiments, we found that using a Large Reasoning Model (LRM) for translation, longer texts are more prone to omissions. When we evaluated the translations generated by LRM and LLM using LLM, we were surprised to find that the LRM results scored higher, which is clearly unreasonable.

As shown in Table \ref{tab_llm_as_judge}, we analyzed the results of DeepSeek V3 and DeepSeek R1. The result of R1 is more concise but has obvious omissions. Compared with the reference translation of 879 tokens, the result of R1 only has 525 tokens. In contrast, the length of the V3 result is close to that of the reference translation. However, when using LLM-as-a-judge for scoring, the result of R1 surprisingly obtained a higher score of 8.5 compared to the 8 points of the V3 result. This finding indicates that the LLM-as-a-judge method may have inaccuracies and biases in evaluating translation quality, especially when dealing with translation results of different lengths.

\begin{table}
\caption{An example of LLM-as-a-judge}
\centering
\label{tab_llm_as_judge}
\begin{tabular}{b{1.4cm}b{7.5cm}b{1.2cm}b{1.2cm}}
\hline
 &  \multicolumn{1}{c}{Text} & \multicolumn{1}{c}{Tokens} & \multicolumn{1}{c}{Score} \\
\hline
Source & 有一天，洛杉矶时报的专栏作家\ Steve Lopez 走在洛杉矶大街上，听到一曲美妙的乐曲。音乐是从一个男人，一个非裔男人，一个颇有魅力，粗旷的流浪汉，用一把只剩下两根弦的小提琴在演奏 ... 而我会永远为他创造音乐，无论是在迪斯尼的音乐厅，或者Skid Row的大街上。 因为他提示着我，为什么我要做音乐家。谢谢 Bruno Giussani:谢谢，谢谢 Robert Gupta. Robert Gupta：现在我给大家演奏一段我厚颜从大提琴手那里窃来的曲子。所以请原谅我 & \multicolumn{1}{c}{1734} & -\\
\hline
Reference & One day, Los Angeles Times columnist Steve Lopez was walking along the streets of downtown Los Angeles when he heard beautiful music. And the source was a man, an African-American man, charming, rugged, homeless, playing a violin that only had two strings. ... And I will always make music with Nathaniel, whether we're at Walt Disney Concert Hall or on Skid Row, because he reminds me why I became a musician. Thank you. Bruno Giussani: Thank you. Thanks. Robert Gupta. Robert Gupta: I'm going to play something that I shamelessly stole from cellists. So, please forgive me. & \multicolumn{1}{c}{879} & - \\
\hline
DeepSeek V3 Out & One day, Steve Lopez, a columnist for the Los Angeles Times, was walking down the streets of Los Angeles when he heard a beautiful melody. The music was being played by a man, an African-American man, a charismatic yet rugged homeless person, using a violin with only two strings left. ... And I will always create music for him, whether at the Disney Concert Hall or on the streets of Skid Row, because he reminds me why I became a musician. Thank you. Bruno Giussani: Thank you, thank you, Robert Gupta. Robert Gupta: Now I will play a piece that I shamelessly stole from a cellist, so please forgive me.
 & \multicolumn{1}{c}{802} & \multicolumn{1}{c}{8} \\
\hline
DeepSeek R1 Out & One day, Los Angeles Times columnist Steve Lopez was walking downtown when he heard exquisite music. It came from a man - an African American man, a charismatic, rough-hewn homeless person - playing a violin with only two strings left. ... I'll always make music with him - whether at Disney Hall or on Skid Row - for he reminds me why I became a musician. Thank you. Bruno Giussani: Thank you, Robert Gupta. Robert Gupta: Now I'll play something I shamelessly stole from the cellists - please forgive me. & \multicolumn{1}{c}{525} & \multicolumn{1}{c}{8.5} \\
\hline
\end{tabular}
\end{table}

\subsubsection{Lack of Interpretability}
The scoring of the LLM-as-a-judge method is inexplicable. We cannot clearly understand the scoring criteria or why a particular score is given. Although the latest Large Reasoning Model has detailed reasoning in the scoring process, the method still has significant interpretability issues overall.

\subsection{Discrepancy Between Metrics and Human Evaluation}

Neither \textit{d}-Comet nor LLM-as-a-judge can fully align with human evaluation. Although these automatic evaluation metrics can provide valuable feedback in some aspects, they still cannot replace human evaluation. Human evaluation can offer more detailed and comprehensive feedback, especially when dealing with complex semantics and context. Therefore, despite significant progress in the development of automatic evaluation metrics, human evaluation remains indispensable.

\section{Future Trends}

\subsection{More User-friendly Document-level Evaluation Methods}

In the future, the development of more user-friendly document-level evaluation methods will be one of the key research focuses, with methods similar to \textit{d}-Comet expected to be further improved and optimized. Specific directions may include:

\paragraph{Reducing the Dependency on Sentence-level Information} 
Current document-level evaluation methods often rely on sentence-level alignment information, which has many limitations in practical applications. Future research may explore reducing the dependency on sentence-level information, conducting evaluations directly at the document level to enhance the applicability and flexibility of evaluation methods.

\paragraph{More Automatic Construction and Utilization of Sentence-level Information} 
Another possible direction is the development of more automatic techniques for constructing or utilizing sentence-level information. For example, through automatic sentence segmentation and alignment algorithms, the quality and usability of sentence-level information can be improved, thereby providing a more reliable basis for document-level evaluation.

\subsection{More Robust LLM-as-a-judge Methods}

To overcome the limitations of existing LLM-as-a-judge methods, future research will focus on developing more robust evaluation methods. Specific directions may include:

\paragraph{Multi-level and Multi-granular LLM-as-a-judge} 
Introducing an approach similar to Chain of Thought (CoT) \cite{kojima2023largelanguagemodelszeroshot}, allowing LLMs to not only focus on the final results during the evaluation process but also identify specific types of errors in translations, such as omissions, over-translations, and mistranslations. Scoring based on these error types can make evaluation results more interpretable and accurate.

\paragraph{Specialized LRM for Machine Translation Evaluation} 
Using human evaluation data to construct training datasets that include reasoning processes and scores, training a Large Reasoning Model (LRM) specifically for machine translation evaluation. This model will be able to better understand the key factors of translation quality and provide more accurate and reliable evaluation results.

\section{Conclusion}

This paper has comprehensively explored the automatic evaluation of document-level translation, analyzing the current status of evaluation, the challenges faced, and the future trends. Although existing evaluation metrics can reflect translation quality to some extent, they still have many shortcomings, such as the lack of reference diversity, dependence on sentence-level alignment information, and the bias and lack of interpretability of LLM-based evaluation methods. Future research will focus on developing more user-friendly and robust evaluation methods, reducing the dependency on sentence-level information, and improving the accuracy and reliability of evaluation by introducing multi-level and multi-granular evaluation approaches and specialized model training. These research directions are expected to bring new breakthroughs to the automatic evaluation of document-level translation and promote the further development of machine translation technology.

\bibliographystyle{splncs04}
\bibliography{custom}
\end{CJK*}
\end{document}